\documentclass[conference, final]{IEEEtran}
\IEEEoverridecommandlockouts

\usepackage{cite}
\usepackage{amsmath,amssymb,amsfonts}
\usepackage{algorithmic}
\usepackage{graphicx}
\usepackage{textcomp}

\usepackage[inline]{enumitem}
\usepackage{xcolor}
\usepackage{witharrows}
\usepackage[flushleft]{threeparttable}
\usepackage{tablefootnote}
\usepackage{array}

\newcolumntype{P}[1]{>{\centering\arraybackslash}m{#1}}

\usepackage{multicol}
\usepackage{multirow}
\usepackage{tabulary}
\usepackage{colortbl}
\definecolor{mygray}{gray}{0.90}
\usepackage{makecell}
\usepackage{booktabs}
\usepackage{subcaption}
\usepackage{stmaryrd}
\usepackage{float}
\usepackage{pifont}

\usepackage{arydshln}
\colorlet{mygray}{gray!15!white}

\usepackage[switch,columnwise]{lineno}

\usepackage{url,hyperref,microtype}
\hypersetup{
    colorlinks=true,
    citecolor=blue,
    linkcolor=blue,
    filecolor=magenta,
    urlcolor=black,
}

\def\BibTeX{{\rm B\kern-.05em{\sc i\kern-.025em b}\kern-.08em
    T\kern-.1667em\lower.7ex\hbox{E}\kern-.125emX}}

\begin{document}

\title{A Multi-Scale Temporal Framework with Dynamic Fusion for EEG-Based Emotion Recognition\\
}

\author{

\IEEEauthorblockN{Stefanos Gkikas}
\IEEEauthorblockA{\textit{Honda Research Institute Japan} \\
Wako City, Japan \\
stefanos.gkikas@jp.honda-ri.com}

\and

\IEEEauthorblockN{Yang Guo}
\IEEEauthorblockA{\textit{Faculty of Information Science} \\
\textit{and Engineering} \\
\textit{Ocean University of China}\\
Qingdao, China \\
diw85827@gmail.com}

\and

\IEEEauthorblockN{Guangliang Li}
\IEEEauthorblockA{\textit{Faculty of Information Science} \\
\textit{and Engineering} \\
\textit{Ocean University of China}\\
Qingdao, China \\
guangliangli@ouc.edu.cn}

\and

\IEEEauthorblockN{Raul Fernandez Rojas}
\IEEEauthorblockA{
\textit{BioSIS (Biosensing \& Intelligent Systems) Lab} \\
\textit{Centre for Intelligent Computing and Systems} \\
\textit{University of Canberra} \\
Canberra, Australia \\
raul.fernandezrojas@canberra.edu.au
}

\and

\IEEEauthorblockN{Giorgos Giannakakis}
\IEEEauthorblockA{\textit{Department of Electronic Engineering} \\
\textit{Hellenic Mediterranean University}\\
Chania, Greece \\
ggian@hmu.gr}

\and

\IEEEauthorblockN{Randy Gomez}
\IEEEauthorblockA{\textit{Honda Research Institute Japan} \\
Wako City, Japan \\
r.gomez@jp.honda-ri.com}

}

\maketitle

\begin{abstract}
Mixed emotions represent a clinically relevant but still underexplored target for automatic emotion recognition. EEG provides millisecond-level access to neural activity, yet most EEG pipelines analyze the signal through a single temporal window, thereby fixing the temporal structure available to the model.
This study introduces a multi-scale temporal framework for EEG-based emotion recognition. The EEG waveform is decomposed into windows of one or several durations, processed by a shared attention-based encoder, and integrated through a dynamic fusion module that assigns sample-specific weights across temporal scales. The framework is evaluated under a subject-independent protocol in binary and three-class settings, with the three-class task including the mixed affective category.
The best results are $65.22\%$ for the two-class task and $45.43\%$ for the three-class task. Both are obtained with three-scale dynamic-fusion configurations and remain substantially above the full-signal baseline. The best-performing temporal scales differ between the two tasks.
Dynamic fusion outperforms concatenation in the highest-scoring two-class configuration and slightly exceeds it in the highest-scoring three-class configuration, although these multi-scale settings require substantially more computation than the full-signal baseline.
\end{abstract}



\section{Introduction}

Mixed emotions refer to emotional episodes in which positive and negative affective components occur at the same time \cite{larsen_mcgraw_2011}. They are not adequately described by a single valence axis. Instead, they form a distinct affective category that requires the two poles to be evaluated independently \cite{oh_tong_2022}. Instruments such as the Positive and Negative Affect Schedule \cite{watson_clark_1988} and valence-arousal-dominance ratings \cite{bradley_lang_1994} are commonly used to assess subjective affective experience. However, reconstructing the conflicting components of a mixed affective state after the event is difficult, and retrospective self-report becomes unreliable when affective states must be monitored continuously \cite{aaker_drolet_2008}.

The global burden of mental health conditions is substantial. Estimates attribute an additional $53$ million cases of major depressive disorder and $76$ million cases of anxiety disorder to the COVID-19 pandemic in a single year \cite{santomauro_herrera_2021}. Mixed affective states are clinically important because they preserve the simultaneous presence of opposing affective components, information that single-valence descriptors discard \cite{larsen_mcgraw_2014}. Evidence from psychotherapy populations further shows that concurrent positive and negative affect can precede improvements in psychological well-being, highlighting the value of characterizing mixed emotional experience for mental health monitoring \cite{adler_hershfield_2012}.

Self-report alone is limited in settings that require continuous and unobtrusive monitoring. EEG frequency-band features have been studied across psychiatric conditions such as depression, bipolar disorder, anxiety, and post-traumatic stress disorder \cite{newson_thiagarajan_2019}. Wearable systems that apply machine learning to physiological signals have also shown promise for the passive detection of anxiety and depression \cite{abd_alrazaq_alsaad_2023}. Among physiological modalities, EEG directly measures central nervous system activity with millisecond resolution, capturing affective dynamics that peripheral biosignals cannot express in the same way \cite{alarcao_fonseca_2019}.

Automatic emotion recognition from EEG has advanced considerably with deep learning, especially for discrete affective classification tasks \cite{song_zheng_2020}. Convolutional and recurrent architectures have been applied to spectral and temporal EEG representations \cite{zheng_lu_2015}, while attention-based models have enabled longer-range dependencies to be modeled within EEG sequences \cite{song_zheng_2023}. Affective responses, however, do not evolve at a single temporal resolution. They include both short, transient reactions and sustained dynamics throughout the full stimulus duration. The analysis window therefore remains a central design choice, as it determines the temporal information exposed to the model \cite{alarcao_fonseca_2019}. Mixed emotional states, where positive and negative components co-occur, remain comparatively underexplored from a computational perspective despite their clinical relevance \cite{larsen_mcgraw_2011}.

This study proposes a multi-scale temporal framework for EEG-based emotion recognition. Time-domain EEG waveforms are decomposed into windows of multiple durations, encoded by a shared attention-based model, and integrated through a dynamic fusion mechanism that assigns sample-adaptive weights across temporal scales. The framework is evaluated in binary and three-class classification settings covering positive, negative, and mixed affective categories.

\section{Related Work}
\label{related_work}

Automatic emotion recognition has been studied across physiological and behavioral modalities, including electrodermal activity, photoplethysmography, cardiovascular signals, and facial video \cite{koelstra_muhl_2012}. EEG occupies a particular role in this setting because it records central nervous system activity at millisecond resolution and provides access to affective correlates that peripheral biosignals cannot directly express \cite{alarcao_fonseca_2019}.
Recent work in affective computing has also examined how the organization and fusion of behavioral and physiological information affect recognition performance. Examples include graph-based facial representations for stress recognition, channel-stacked visual representations of EDA for stress detection, and modality-agnostic fusion of heterogeneous physiological signals for cognitive workload assessment~\cite{kassiotis_stressgat_acii_2026, gkikas_eda_stress_prai_2026, gkikas_workload_acii_2026}.

Early EEG-based emotion recognition methods relied on hand-crafted spectral descriptors paired with shallow classifiers. Later approaches introduced deep architectures, including deep belief networks operating on differential entropy features \cite{zheng_lu_2015}. Differential entropy across the delta, theta, alpha, beta, and gamma bands has since become one of the most common feature representations for EEG emotion classification \cite{duan_zhu_2013}. Graph convolutional networks were then introduced to model electrode topology and learn functional relationships between channels beyond static feature representations \cite{song_zheng_2020}. More recently, transformer-based decoders that combine convolutional modules with self-attention have improved the modeling of local spatiotemporal patterns and long-range dependencies \cite{song_zheng_2023}.
Related studies have also explored multimodal, modality-agnostic, and spatiotemporal representation learning across behavioral and physiological signal analysis~\cite{gkikas_tsiknakis_slr_2023,gkikas_tsiknakis_painvit_2024,gkikas_arzate_pain_icmi_2026,gkikas_arzate_eeite_pain_2026,gkikas_tsiknakis_thermal_2024,gkikas_reface_acii_2026}.

Temporal segmentation is a key design factor in EEG-based emotion recognition. The selected analysis window determines the balance between short-term reactivity and longer affective dynamics. Studies on public benchmarks have shown that performance is strongly affected by window length. Wavelet-based features extracted from $3$- to $12$-second segments improved arousal and valence classification compared with full-trial analysis \cite{candra_yuwono_2015}. Similar behavior has been reported for deep networks, where CNN performance under subject-independent evaluation varies with the temporal window and differs between arousal and valence \cite{keelawat_thammasan_2021}. Other studies confirm that the effective temporal resolution depends on both the dataset and the affective dimension, with optimal windows ranging from a few seconds to more than ten seconds \cite{ouyang_zhou_2022}.

Beyond fixed-window analysis, multi-scale architectures have been developed to represent short-term and sustained EEG dynamics within a single model. Multi-scale convolutional and graph-based models encode EEG activity at multiple temporal resolutions \cite{ms_imamba_2024, wang_msgm_2025}, based on the premise that affective processes unfold across multiple temporal scales. A related multi-window fusion strategy has also been explored for pain recognition from respiratory signals, where a single cross-attention transformer aggregates representations extracted at different temporal resolutions \cite{gkikas_kyprakis_resp_2025}.
Unlike that approach, which applies multi-window modeling to a single respiration signal for pain recognition, the present framework operates directly on multichannel EEG waveforms, processes variable-length windows through a shared multi-layer segment-based encoder, and uses sample-adaptive fusion to weight temporal scales and positions for each recording.
Despite these developments, most EEG-based emotion recognition studies still focus on standard discrete or dimensional categories. The co-occurrence of positive and negative affect that characterizes mixed emotional states remains largely unaddressed in computational approaches \cite{larsen_mcgraw_2011}.


\section{Methodology}
\label{sec:methodology}

This section describes the proposed multi-scale temporal framework. Time-domain EEG waveforms are cropped to a common length, decomposed into windows of one or several durations, encoded by a shared attention-based model, and integrated through dynamic fusion. The training objective, augmentation procedure, and regularization settings are also presented.

\subsection{Signal Representation}

Each EEG recording is represented as a multi-channel waveform sampled at a fixed rate $f_s$. A sample consists of a tensor
\begin{equation}
\mathbf{X}_{\mathrm{raw}} \in \mathbb{R}^{C \times L_0},
\end{equation}
where $C$ is the number of EEG channels and $L_0$ is the raw signal length. A stride $r \in \mathbb{N}$ can be optionally applied along the temporal axis, producing a strided signal of length $L_0/r$.
In all experiments reported in this study, $r=1$; therefore, the EEG waveform remains at its original sampling rate of $300$~Hz.
In this work, no hand-crafted feature extraction is applied; the model operates directly on the waveform. All signals are cropped to a fixed target length $L$ selected to match the shortest recording available in the dataset:
\begin{equation}
L = \left\lfloor L_{\min}/r \right\rfloor.
\end{equation}
During training, the crop position is uniformly sampled from the valid range $[0, L_0/r-L]$, yielding multiple temporal views per epoch for recordings longer than $L$. During validation and testing, the crop position is fixed at zero, yielding a deterministic first-$L$ crop for every sample. The resulting tensor is
\begin{equation}
\mathbf{X} \in \mathbb{R}^{C \times L}.
\end{equation}

\subsection{Multi-Scale Temporal Windowing}

Given the cropped signal $\mathbf{X} \in \mathbb{R}^{C \times L}$, a set of window durations is defined as
\begin{equation}
\mathcal{W} = \{w_1, w_2, \ldots, w_K\}, \quad
w_k \in \{\mathrm{full}, 2, 5, 10\}\ \text{seconds},
\end{equation}
where $\mathrm{full}$ denotes the entire cropped signal used as a single window. Each duration $w_k$ produces $n_k$ non-overlapping windows of length $\ell_k = w_k f_s/r$ samples,
\begin{equation}
n_k = \left\lfloor L/\ell_k \right\rfloor,
\end{equation}
so that the total number of windows extracted per sample is
\begin{equation}
N = \sum_{k=1}^{K} n_k.
\end{equation}
For $w_k=\mathrm{full}$, $\ell_k=L$ and $n_k=1$. Each window is a tensor
\begin{equation}
\mathbf{X}^{(k,i)} \in \mathbb{R}^{C \times \ell_k},
\quad i=1,\ldots,n_k,
\end{equation}
containing all $C$ EEG channels within its temporal interval.

\subsection{Shared Attention-Based Encoder}
\label{sec:encoder}

All $N$ windows are processed by a single shared encoder. Weight sharing across scales enables the model to learn a common representation function regardless of window duration, while keeping the encoder parameter count independent of $N$; only the lightweight fusion head and the window-specific normalization layers vary with the window configuration. Before encoding, each window is normalized through a window-length-specific layer normalization module.

\subsubsection{Tokenization}

Each window $\mathbf{X}^{(k,i)}$ is treated as a 1D signal with $C$ channels and $\ell_k$ temporal positions. The temporal axis is flattened into a sequence of $\ell_k$ tokens, each carrying the $C$ channel values at the corresponding time step. Geometric information is incorporated by encoding each normalized time position $t \in [-1,1]$ with Fourier features:
\begin{equation}
\begin{split}
\gamma(t)=\bigl[&
\sin(\pi s_1t),\ \cos(\pi s_1t),\ \ldots,\\
&\sin(\pi s_Mt),\ \cos(\pi s_Mt),\ t
\bigr],
\end{split}
\end{equation}
where $\{s_m\}_{m=1}^{M}$ span $[1,f_{\max}/2]$. The model uses $M=6$ frequency bands and $f_{\max}=10$, adding $2M+1=13$ positional features per token. Channel values and positional features are concatenated at each step to form the token matrix
\begin{equation}
\mathbf{T}^{(k,i)}
\in \mathbb{R}^{\ell_k \times C'},
\quad C'=C+(2M+1).
\end{equation}

\subsubsection{Segmentation}

The token sequence is partitioned into $S=100$ contiguous segments of length $n_s=\lceil \ell_k/S \rceil$, padded when necessary. The tokens of segment $s$ are denoted
$\tilde{\mathbf{T}}^{(k,i)}_s \in \mathbb{R}^{n_s \times C'}$.
With $r=1$, the $2$-, $5$-, $10$-, and $20$-second inputs contain $600$, $1500$, $3000$, and $6000$ samples, respectively. Using $S=100$ therefore produces segments of $6$, $15$, $30$, and $60$ samples, corresponding to local intervals of $20$, $50$, $100$, and $200$~ms. Cross-attention performs learned aggregation within each interval rather than fixed averaging, while self-attention across the resulting $100$ segment states models dependencies over the complete window.

\subsubsection{Hierarchical block schedule}

The encoder consists of $D=6$ layers with progressively compressed latent representations. The per-layer latent dimensionality is
\begin{equation}
d_\ell \in \{128,\ 128,\ 96,\ 96,\ 64,\ 32\},
\quad \ell=1,\ldots,6,
\end{equation}
while the number of segment states is fixed at $S=100$ across all layers. Cross-attention uses $\{8,8,6,6,4,2\}$ heads across the six layers with a head dimension of $16$ in every layer. Self-attention uses the same head schedule and head dimension. The resulting attention inner dimensionalities are $\{128,128,96,96,64,32\}$. Each cross-attention layer is followed by $R_\ell \in \{8,8,4,4,2,1\}$ self-attention blocks. Attention and feedforward dropout are fixed at $0.20$ across all layers.

\subsubsection{Segment states}

A single latent seed is derived at runtime from a set of $M_0$ learnable global parameters $\{\boldsymbol{\ell}_m\}_{m=1}^{M_0}$:
\begin{equation}
\boldsymbol{\ell}_{\mathrm{init}}
=
\frac{1}{M_0}
\sum_{m=1}^{M_0}
\boldsymbol{\ell}_m
\in \mathbb{R}^{d_1},
\end{equation}
and replicated across all $S$ segments to instantiate the initial segment states. Segment-specific representations are not learned directly; they emerge through the attention updates.

\subsubsection{Asymmetric cross-attention}

At each layer $\ell$, every segment state aggregates information exclusively from its corresponding token subset:
\begin{equation}
\mathbf{u}_s^{(\ell)}
=
\mathbf{u}_s^{(\ell-1)}
+
\mathrm{Attn}
\bigl(
\mathbf{u}_s^{(\ell-1)},
\tilde{\mathbf{T}}^{(k,i)}_s
\bigr),
\end{equation}
where $\mathbf{u}_s^{(\ell-1)} \in \mathbb{R}^{1 \times d_\ell}$ provides the query and $\tilde{\mathbf{T}}^{(k,i)}_s \in \mathbb{R}^{n_s \times C'}$ provides the keys and values. The operation is asymmetric: the query side is a single vector while the key-value side spans $n_s \gg 1$ tokens of dimension $C'$, and the resulting attention matrix is $1 \times n_s$ rather than square. All $S$ segments are processed in parallel by packing them into the batch dimension.

\subsubsection{Self-attention across segments}

After cross-attention, the segment states are stacked as
$\mathbf{U}^{(\ell)} \in \mathbb{R}^{S \times d_\ell}$ and refined through $R_\ell$ self-attention blocks with pre-layer normalization and residual connections:
\begin{equation}
\mathbf{U}^{(\ell)}
\leftarrow
\mathbf{U}^{(\ell)}
+
\mathrm{Attn}
\bigl(
\mathbf{U}^{(\ell)},
\mathbf{U}^{(\ell)}
\bigr).
\end{equation}
Between consecutive layers, the segment representations are projected from $d_\ell$ to $d_{\ell+1}$ through a linear transformation when the dimensionality changes.

\subsubsection{Window embedding}

After the final layer, the $S$ segment states are averaged across the segment axis to produce a single embedding per window:
\begin{equation}
\mathbf{e}^{(k,i)}
=
\frac{1}{S}
\sum_{s=1}^{S}
\mathbf{U}^{(D)}_s
\in \mathbb{R}^{d},
\end{equation}
where $d=d_D=32$. All windows are represented in the same embedding space $\mathbb{R}^{d}$ independently of their duration, since the encoder is shared across scales.

\subsection{Dynamic Fusion of Window Embeddings}
\label{sec:fusion}

The $N$ window embeddings $\{\mathbf{e}^{(k,i)}\}$ produced by the shared encoder are collected into a matrix
\begin{equation}
\mathbf{Z}
=
[\mathbf{e}_1;\,\mathbf{e}_2;\,\ldots;\,\mathbf{e}_N]
\in \mathbb{R}^{N \times d}
\end{equation}
and integrated by a fusion module to produce a single sample-level representation for classification. Two fusion strategies are considered.

\subsubsection{Concatenation}

The embeddings are concatenated along the channel dimension and passed through a two-layer MLP with layer normalization, GELU activation, and dropout:
\begin{equation}
\mathbf{E}
=
\mathrm{concat}
\bigl(
\mathbf{e}^{(1,1)},
\ldots,
\mathbf{e}^{(K,n_K)}
\bigr)
\in \mathbb{R}^{Nd},
\end{equation}
\begin{equation}
\hat{\mathbf{y}}=\mathrm{MLP}(\mathbf{E}).
\end{equation}
Under this scheme, the model has no explicit sample-adaptive weighting before the classification head; the window ordering is fixed for every sample.

\subsubsection{Dynamic fusion}

\begin{figure*}
\begin{center}
\includegraphics[scale=0.55]{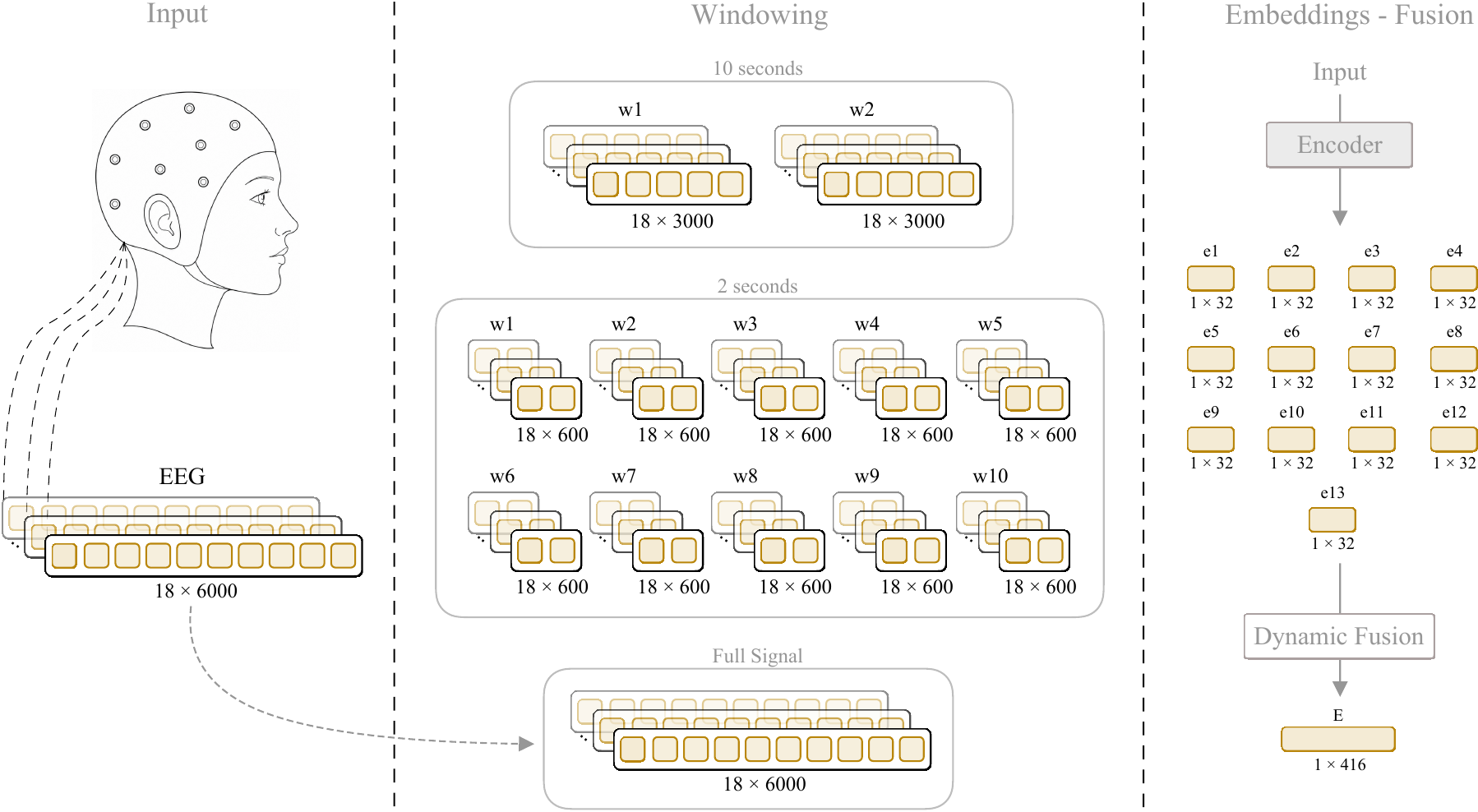}
\end{center}
\caption{Overview of the proposed multi-scale temporal framework, illustrated with the $\{\mathrm{full},2,10\}$ configuration. The $18$-channel EEG signal of length $6000$ samples is decomposed into three temporal scales: one full-signal window ($18 \times 6000$), two $10$-second windows ($18 \times 3000$ each), and ten $2$-second windows ($18 \times 600$ each), for a total of $N=13$ windows. All windows are processed by the same shared encoder into $32$-dimensional embeddings $\mathbf{e}_1,\ldots,\mathbf{e}_{13}$. The embeddings are integrated by the dynamic fusion module into a single sample-level representation $\mathbf{E}\in\mathbb{R}^{416}$, which is passed to the classification head.}
\label{overview}
\end{figure*}

A learnable query vector $\mathbf{q}\in\mathbb{R}^{d}$ and a set of learnable per-slot position embeddings
$\{\mathbf{p}_i\}_{i=1}^{N}$,
$\mathbf{p}_i\in\mathbb{R}^{d}$ are introduced. Attention scores are computed by shifting each embedding with its position and taking a scaled dot product with the query:
\begin{equation}
\alpha_i
=
\frac{
(\mathbf{e}_i+\mathbf{p}_i)^{\top}\mathbf{q}
}{
\sqrt{d}
},
\quad i=1,\ldots,N,
\end{equation}
and normalized across the $N$ windows through a softmax:
\begin{equation}
w_i
=
\frac{
\exp(\alpha_i)
}{
\sum_{j=1}^{N}\exp(\alpha_j)
}.
\end{equation}
The weighted window embeddings are then concatenated to form the fused representation:
\begin{equation}
\mathbf{E}
=
\mathrm{concat}
\bigl(
w_1\mathbf{e}_1,
w_2\mathbf{e}_2,
\ldots,
w_N\mathbf{e}_N
\bigr)
\in \mathbb{R}^{Nd}.
\end{equation}
$\mathbf{E}$ is passed through the same MLP head as in the concatenation strategy. The weights $\{w_i\}$ are sample-specific, so the model can assign different importance to different temporal scales and positions depending on the content of each recording.
An overview of the framework, for the $\{\mathrm{full},2,10\}$ configuration, is shown in Fig.~\ref{overview}.

\subsection{Training and Augmentation}

The model is trained for $200$ epochs using the AdamW optimizer with a base learning rate of $10^{-4}$, a weight decay of $0.05$, and a cosine learning-rate schedule with a $20$-epoch warmup from $10^{-6}$ and a minimum learning rate of $10^{-5}$. The batch size is $32$ for all splits. Mixed-precision training is used throughout. All experiments were conducted with a fixed random seed of $3407$ using \texttt{seed\_everything}. Deterministic cuDNN execution was enabled, while cuDNN benchmarking was disabled.

\subsubsection{Loss}

The three-class task uses weighted cross-entropy with class weights $[0.83,1.18,1.09]$ for the mixed, negative, and positive classes, respectively, applied to account for the class imbalance in the training set. The two-class task uses standard cross-entropy, excluding the mixed class. Label smoothing with a smoothing rate of $0.15$ is applied during training and disabled during evaluation.

\subsubsection{Augmentation}

All augmentations operate on the full cropped signal $\mathbf{X}\in\mathbb{R}^{C\times L}$ prior to window extraction, with synchronized decisions across all channels to preserve the inter-channel amplitude relationships and temporal synchrony that carry affective information. Two augmentations are applied stochastically during training. Additive noise draws a per-sample decision with probability $p_n\sim\mathcal{U}(0.50,0.80)$ and a signal-to-noise scaling factor $\phi\sim\mathcal{U}(1,1000)$; independent noise realizations are generated per channel at the same signal-to-noise ratio. Temporal masking draws a per-sample decision with probability $p_m\sim\mathcal{U}(0.50,0.80)$ and a masked fraction $\rho\sim\mathcal{U}(0.15,0.30)$; the same temporal mask is broadcast to all $C$ channels, with the masked interval placed at the start, end, or center of the signal with equal probability.

\subsubsection{Regularization}

Dropout with probability $0.20$ is applied in the attention and feedforward sublayers of every encoder block, and the fusion head uses a dropout of $0.20$ before the final linear projection.


\section{Experimental Evaluation \& Results}

The proposed framework is evaluated on two tasks: binary classification of affective states as positive or negative and three-way classification as positive, negative, or mixed. Validation performance is reported using balanced accuracy, macro-averaged precision, and macro-averaged F1 score. Test performance is reported using balanced accuracy. The test set is used only for final reporting; model checkpoints and hyperparameters are selected on the validation set, and results are reported for all evaluated configurations.

\subsection{Data Collection}
\label{ssec:data_collection}

The experiments use the DMER dataset \cite{yang_liu_2024}. The dataset includes $73$ participants aged $18$ to $35$ (mean age $23.06$, SD $3.37$); data from $7$ of the original $80$ participants were excluded because of physiological signal quality issues. Participants watched $32$ short video clips selected from the Stanford film library through rule-based filtering and expert evaluation. The clips were designed to elicit positive, negative, and mixed emotional states: $8$ clips for positive emotions, $8$ for negative emotions, and $16$ for mixed emotions. Mixed emotion is defined as the simultaneous co-occurrence of positive and negative affect and was validated through PANAS and valence-arousal-dominance self-reports. EEG was recorded with the DSI-24 wireless dry-electrode system at $300$~Hz across $21$ channels following the international $10$--$20$ system.

This study uses EEG as the input modality. The time-aligned preprocessed EEG signals provided by DMER are used directly. The dataset authors applied independent component analysis, bandpass filtering from $1$ to $50$~Hz, $50$~Hz notch filtering, baseline removal, and re-referencing to Pz with ear electrodes excluded, resulting in $18$ channels per trial. Recordings have two durations, $20$ seconds and $30$ seconds, and these lengths are not uniformly distributed across affective categories. To prevent duration from becoming a predictive cue, all recordings are cropped to $20$ seconds. During training, the crop position is sampled uniformly from the available range. During validation and testing, the first $20$ seconds are used. The three-class task classifies affective states as positive, negative, or mixed. The two-class task distinguishes positive from negative affective states, with the mixed class excluded.

Subjects are split at the participant level, so no individual appears in more than one set. To reduce performance inflation due to subject-difficulty imbalance, a stratified split protocol is used. Leave-one-subject-out cross-validation under the three-class task first estimates per-subject classification difficulty. Subjects are then ranked by their combined z-score and assigned to four quartile-based groups. This ranking is used only to construct a balanced subject partition and does not influence model training, checkpoint selection, or hyperparameter tuning. The final split contains $48$ training, $12$ validation, and $13$ testing subjects, with each set containing subjects from all four difficulty groups. The exact partition is reported in Table~\ref{tab:subject_split} for reproducibility.

\begin{table*}
\caption{Subject-level split by difficulty group. Subjects are ranked by combined z-score and assigned to four quartile-based groups (Q1\,=\,hardest, Q4\,=\,easiest).}
\label{tab:subject_split}

\begin{center}
\begin{threeparttable}
\begin{tabular}{P{1.8cm} P{3.5cm} P{3.5cm} P{3.5cm} P{3.5cm}}
\toprule

\multirow[c]{3}{*}{Split} &
\multicolumn{4}{c}{Difficulty Group} \\

\cmidrule(lr){2-5}
& Q1 -- Hard & Q2 -- Med-Hard & Q3 -- Med-Easy & Q4 -- Easy \\

\midrule
\midrule
Training (48) &
8, 12, 20, 25, 36, 37, 43, 45, 56, 60, 66, 71 &
14, 30, 33, 35, 39, 50, 51, 53, 55, 65, 72, 73 &
5, 6, 9, 10, 15, 23, 32, 52, 54, 59, 70, 78 &
11, 24, 26, 34, 40, 42, 49, 61, 63, 64, 68, 76 \\\hdashline

Validation (12) &
19, 48, 58 &
22, 41, 79 &
29, 47, 77 &
62, 67, 69 \\\hdashline

Testing (13) &
28, 74, 75, 80 &
21, 38, 57 &
1, 2, 7 &
18, 44, 46 \\

\bottomrule
\end{tabular}

\begin{tablenotes}[para,flushleft]
\scriptsize
\item Q1: $z<-0.36$;\quad
Q2: $-0.36\leq z<-0.08$;\quad
Q3: $-0.08\leq z<0.17$;\quad
Q4: $z\geq0.17$.
\end{tablenotes}

\end{threeparttable}
\end{center}
\end{table*}


\subsection{Two-Class Emotion Recognition}
\label{sec:2class}

Table~\ref{table:2class} reports the results for the two-class task across all window configurations and fusion strategies. Test accuracy ranges from $56.64\%$ to $65.22\%$ across the $29$ evaluated configurations. The full-signal baseline, which processes the entire cropped signal as a single window without fusion, reaches $57.41\%$ test accuracy. Most windowed configurations improve on this baseline, indicating that shorter temporal segments provide discriminative information not fully captured by the full-signal representation.

For single-scale settings, the strongest results appear at the $2$-second scale for both concatenation and dynamic fusion ($63.75\%$ and $63.26\%$, respectively). The $5$-second scale also performs well under concatenation ($63.76\%$), while dynamic fusion reaches $60.40\%$. At the $10$-second scale, concatenation achieves $60.33\%$ and dynamic fusion reaches $61.27\%$. Combining multiple temporal scales further improves performance in several cases: concatenation reaches $62.75\%$ with $\{\mathrm{full},2,10\}$, and dynamic fusion reaches $64.20\%$ with $\{\mathrm{full},10\}$.

The best results are obtained with dynamic fusion on three-scale configurations that combine the full signal with short and longer windows. The $\{\mathrm{full},2,5\}$ configuration reaches $65.19\%$, while $\{\mathrm{full},2,10\}$ reaches $65.22\%$, the highest accuracy in the two-class task. The corresponding concatenation configurations remain lower, at $59.84\%$ and $62.75\%$. Adding all four scales $\{\mathrm{full},2,5,10\}$ does not improve performance, reaching $57.49\%$ with concatenation and $61.82\%$ with dynamic fusion. Dynamic fusion is therefore most useful when applied to a compact set of complementary temporal scales rather than to the full redundant scale grid.

\begin{table}
\caption{Two-class classification performance (positive vs.\ negative) across all window configurations and fusion strategies.}
\label{table:2class}

\begin{center}
\begin{threeparttable}
\begin{tabular}{P{1.8cm} P{1.1cm} P{0.9cm} P{0.9cm} P{0.4cm} P{1.2cm}}
\toprule

\multicolumn{2}{c}{Input} &
\multicolumn{3}{c}{Validation} &
\multicolumn{1}{c}{Testing} \\

\cmidrule(lr){1-2}
\cmidrule(lr){3-5}
\cmidrule(lr){6-6}

Windows (s) & Fusion & Accuracy & Precision & F1 & Accuracy \\

\midrule
\midrule
full            & --      & 55.73 & 55.81 & 55.58 & 57.41 \\\midrule

2               & concat  & 58.33 & 58.35 & 58.32 & 63.75 \\\hdashline
5               & concat  & 59.90 & 59.91 & 59.89 & 63.76 \\\hdashline
10              & concat  & 56.77 & 56.99 & 56.43 & 60.33 \\\hdashline
full, 2         & concat  & 55.73 & 55.91 & 55.38 & 60.87 \\\hdashline
full, 5         & concat  & 57.29 & 58.20 & 56.07 & 62.29 \\\hdashline
full, 10        & concat  & 56.25 & 56.25 & 56.25 & 57.39 \\\hdashline
2, 5            & concat  & 58.85 & 58.86 & 58.85 & 60.89 \\\hdashline
2, 10           & concat  & 58.85 & 58.86 & 58.84 & 62.35 \\\hdashline
5, 10           & concat  & 61.46 & 61.58 & 61.35 & 56.64 \\\hdashline
full, 2, 5      & concat  & 55.21 & 55.62 & 54.37 & 59.84 \\\hdashline
full, 2, 10     & concat  & 58.33 & 58.51 & 58.11 & 62.75 \\\hdashline
full, 5, 10     & concat  & 61.46 & 61.79 & 61.19 & 57.93 \\\hdashline
2, 5, 10        & concat  & 62.50 & 62.86 & 62.24 & 58.41 \\\hdashline
full, 2, 5, 10  & concat  & 56.25 & 56.27 & 56.21 & 57.49 \\\midrule

2               & dynamic & 58.33 & 58.42 & 58.22 & 63.26 \\\hdashline
5               & dynamic & 61.98 & 63.58 & 60.82 & 60.40 \\\hdashline
10              & dynamic & 56.77 & 56.78 & 56.76 & 61.27 \\\hdashline
full, 2         & dynamic & 56.77 & 57.05 & 56.34 & 60.78 \\\hdashline
full, 5         & dynamic & 60.42 & 61.11 & 59.79 & 60.83 \\\hdashline
full, 10        & dynamic & 56.77 & 63.20 & 50.78 & 64.20 \\\hdashline
2, 5            & dynamic & 60.42 & 60.58 & 60.26 & 57.42 \\\hdashline
2, 10           & dynamic & 60.42 & 60.49 & 60.35 & 62.75 \\\hdashline
5, 10           & dynamic & 55.21 & 55.56 & 54.50 & 61.80 \\\hdashline
full, 2, 5      & dynamic & 55.73 & 55.84 & 55.53 & 65.19 \\\hdashline
full, 2, 10     & dynamic & 58.33 & 58.71 & 57.88 & \textbf{65.22} \\\hdashline
full, 5, 10     & dynamic & 59.38 & 59.58 & 59.16 & 57.85 \\\hdashline
2, 5, 10        & dynamic & 57.81 & 57.92 & 57.67 & 60.85 \\\hdashline
full, 2, 5, 10  & dynamic & 60.42 & 60.49 & 60.35 & 61.82 \\

\bottomrule
\end{tabular}

\begin{tablenotes}[para,flushleft]
\scriptsize
\item ``full'' denotes the entire $20$-second cropped signal used as a single window. Integer entries denote the window duration in seconds. \textbf{Bold} marks the peak test accuracy.
\end{tablenotes}

\end{threeparttable}
\end{center}
\end{table}


\subsection{Three-Class Emotion Recognition}
\label{sec:3class}

Table~\ref{table:3class} reports the results for the three-class task, which includes the mixed affective category together with positive and negative. Test accuracy ranges from $37.22\%$ to $45.43\%$ across the $29$ configurations. The full-signal baseline reaches $37.24\%$, above the three-class chance level of $33.33\%$ but among the lowest results in the evaluation. As in the two-class task, most windowed configurations outperform the full-signal baseline.

Among single-scale configurations, the $2$-second window gives the strongest single-scale result in both fusion families, reaching $44.93\%$ under concatenation and $41.08\%$ under dynamic fusion. The $5$- and $10$-second scales reach $39.51\%$ and $38.05\%$ under concatenation, and $42.24\%$ and $38.61\%$ under dynamic fusion. Two-scale combinations provide additional gains in selected cases, with $\{2,10\}$ reaching $45.28\%$ under dynamic fusion and $\{2,5\}$ reaching $43.05\%$ under concatenation.

The best three-class results are obtained by combining short, medium, and long windows without the full signal. The $\{2,5,10\}$ configuration reaches $45.31\%$ under concatenation and $45.43\%$ under dynamic fusion, the highest test accuracy for this task. Unlike the two-class setting, the highest-scoring three-class configurations do not include the full-signal window. Under the present DMER evaluation, this suggests a stronger role for temporally local information, possibly because positive and negative affective components emerge within shorter or partially overlapping intervals.

\begin{table}
\caption{Three-class classification performance (positive, negative, mixed) across all window configurations and fusion strategies.}
\label{table:3class}

\begin{center}
\begin{threeparttable}
\begin{tabular}{P{1.8cm} P{1.1cm} P{0.9cm} P{0.9cm} P{0.4cm} P{1.2cm}}
\toprule

\multicolumn{2}{c}{Input} &
\multicolumn{3}{c}{Validation} &
\multicolumn{1}{c}{Testing} \\

\cmidrule(lr){1-2}
\cmidrule(lr){3-5}
\cmidrule(lr){6-6}

Windows (s) & Fusion & Accuracy & Precision & F1 & Accuracy \\

\midrule
\midrule
full            & --      & 33.51 & 25.15 & 28.34 & 37.24 \\\midrule

2               & concat  & 44.97 & 49.28 & 44.03 & 44.93 \\\hdashline
5               & concat  & 38.54 & 35.18 & 34.14 & 39.51 \\\hdashline
10              & concat  & 38.72 & 30.92 & 33.39 & 38.05 \\\hdashline
full, 2         & concat  & 43.06 & 43.51 & 41.01 & 43.33 \\\hdashline
full, 5         & concat  & 44.27 & 45.69 & 42.64 & 41.25 \\\hdashline
full, 10        & concat  & 34.72 & 34.75 & 33.41 & 37.22 \\\hdashline
2, 5            & concat  & 42.36 & 52.25 & 39.27 & 43.05 \\\hdashline
2, 10           & concat  & 42.71 & 44.75 & 39.62 & 40.78 \\\hdashline
5, 10           & concat  & 44.97 & 47.14 & 43.37 & 40.15 \\\hdashline
full, 2, 5      & concat  & 42.01 & 41.10 & 39.91 & 43.99 \\\hdashline
full, 2, 10     & concat  & 39.58 & 39.66 & 38.69 & 42.04 \\\hdashline
full, 5, 10     & concat  & 43.40 & 44.62 & 42.10 & 41.72 \\\hdashline
2, 5, 10        & concat  & 41.84 & 41.17 & 38.86 & 45.31 \\\hdashline
full, 2, 5, 10  & concat  & 39.24 & 39.53 & 38.05 & 42.55 \\\midrule

2               & dynamic & 42.88 & 44.14 & 43.01 & 41.08 \\\hdashline
5               & dynamic & 41.32 & 32.06 & 35.76 & 42.24 \\\hdashline
10              & dynamic & 41.15 & 42.22 & 40.57 & 38.61 \\\hdashline
full, 2         & dynamic & 38.89 & 43.73 & 35.17 & 40.31 \\\hdashline
full, 5         & dynamic & 42.71 & 41.90 & 41.65 & 42.88 \\\hdashline
full, 10        & dynamic & 36.81 & 38.93 & 36.37 & 38.34 \\\hdashline
2, 5            & dynamic & 42.53 & 43.58 & 42.67 & 41.75 \\\hdashline
2, 10           & dynamic & 41.84 & 44.61 & 41.15 & 45.28 \\\hdashline
5, 10           & dynamic & 40.80 & 44.50 & 37.57 & 40.78 \\\hdashline
full, 2, 5      & dynamic & 40.97 & 41.22 & 39.98 & 42.21 \\\hdashline
full, 2, 10     & dynamic & 40.63 & 43.19 & 37.59 & 40.95 \\\hdashline
full, 5, 10     & dynamic & 40.80 & 41.46 & 39.77 & 40.29 \\\hdashline
2, 5, 10        & dynamic & 44.62 & 46.27 & 42.81 & \textbf{45.43} \\\hdashline
full, 2, 5, 10  & dynamic & 42.19 & 43.46 & 41.02 & 44.45 \\

\bottomrule
\end{tabular}

\begin{tablenotes}[para,flushleft]
\scriptsize
\item ``full'' denotes the entire $20$-second cropped signal used as a single window. Integer entries denote the window duration in seconds. \textbf{Bold} marks the peak test accuracy.
\end{tablenotes}

\end{threeparttable}
\end{center}
\end{table}


\subsection{Computational Cost}
\label{sec:cost}

Table~\ref{table:cost} reports the computational cost for the $15$ unique window configurations. The parameter count remains between $4.81$ and $4.97$~M, reflecting the shared-encoder design. The same encoder processes every window, so the main encoder parameter count is independent of the number of extracted windows. The small variation across configurations comes from the fusion head, whose input dimensionality scales with the total number of windows $N$, and from the window-length-specific normalization layers.

\begin{figure*}
\begin{center}
\includegraphics[scale=0.65]{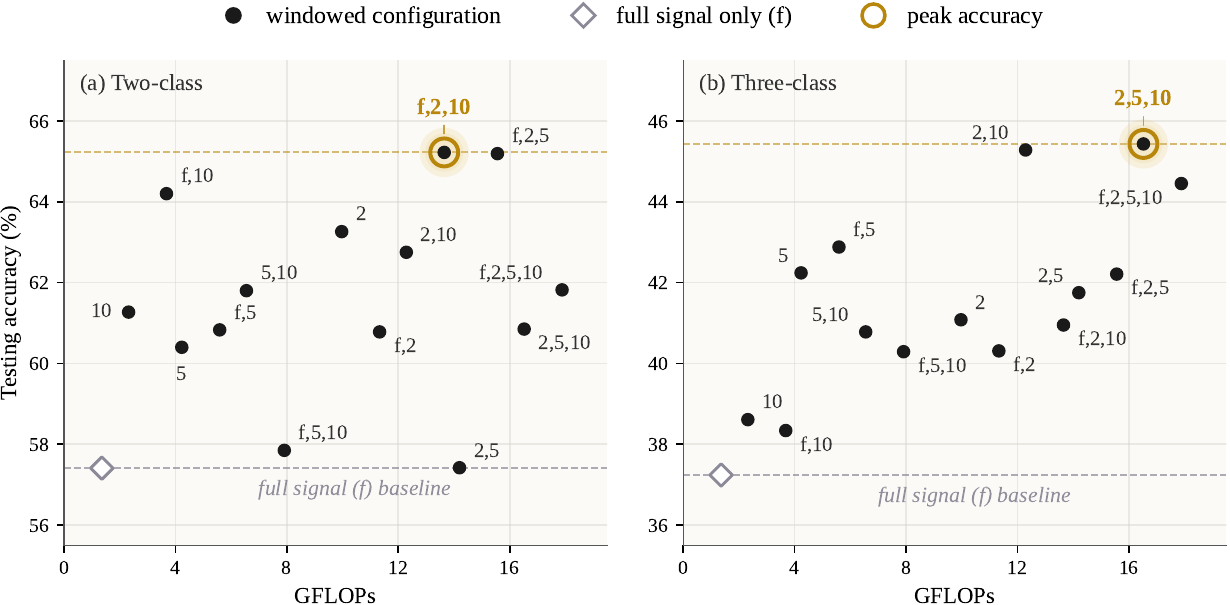}
\end{center}
\caption{Test accuracy versus computational cost across all $15$ window configurations under dynamic fusion, for the two-class task (a) and the three-class task (b). Each marker denotes a single window configuration; the peak-accuracy configuration in each task is highlighted. The dashed horizontal line marks the full-signal baseline. The peak configurations, $\{\mathrm{full},2,10\}$ for the two-class task and $\{2,5,10\}$ for the three-class task, remain below the maximum-cost four-scale setting, although both lie toward the upper end of the evaluated computational range.}
\label{cost}
\end{figure*}

GFLOPs vary more strongly with the window configuration, from $1.36$~GFLOPs for the full-signal baseline to $17.88$~GFLOPs for the four-scale setting. Configurations with several short windows are the most expensive because the number of encoded window instances increases with $N$, even though windows of the same duration are batched within the shared encoder. The highest-scoring configurations do not use the maximum-cost four-scale setting, although both remain computationally demanding. The $\{\mathrm{full},2,10\}$ configuration requires $13.65$~GFLOPs, while the $\{2,5,10\}$ configuration requires $16.52$~GFLOPs, the second-highest cost among the evaluated settings. Fig.~\ref{cost} shows the accuracy-cost trade-off for all $15$ configurations under dynamic fusion.

\begin{table}
\caption{Computational cost across the $15$ unique window configurations.}
\label{table:cost}

\begin{center}
\begin{threeparttable}
\begin{tabular}{P{2.4cm} P{1.6cm} P{1.4cm}}
\toprule

Windows (s) & Params (M) & GFLOPs \\

\midrule
\midrule
full            & 4.82 & 1.36  \\\hdashline
2               & 4.86 & 9.97  \\\hdashline
5               & 4.81 & 4.23  \\\hdashline
10              & 4.81 & 2.32  \\\hdashline
full, 2         & 4.88 & 11.33 \\\hdashline
full, 5         & 4.83 & 5.59  \\\hdashline
full, 10        & 4.82 & 3.68  \\\hdashline
2, 5            & 4.91 & 14.20 \\\hdashline
2, 10           & 4.88 & 12.29 \\\hdashline
5, 10           & 4.83 & 6.55  \\\hdashline
full, 2, 5      & 4.94 & 15.56 \\\hdashline
full, 2, 10     & 4.91 & 13.65 \\\hdashline
full, 5, 10     & 4.85 & 7.91  \\\hdashline
2, 5, 10        & 4.95 & 16.52 \\\hdashline
full, 2, 5, 10  & 4.97 & 17.88 \\

\bottomrule
\end{tabular}

\begin{tablenotes}[para,flushleft]
\scriptsize
\item ``full'' denotes the entire $20$-second signal used as a single window. Integer entries denote the window duration in seconds.
\end{tablenotes}

\end{threeparttable}
\end{center}
\end{table}


\subsection{Overall Analysis \& Discussion}

The proposed multi-scale temporal framework reaches its best results when multiple temporal scales are combined through dynamic fusion. The peak test accuracies are $65.22\%$ for the two-class task and $45.43\%$ for the three-class task. Both peaks come from three-scale configurations that pair one short window ($2$~seconds) with one or two longer windows, rather than from the finest or coarsest temporal resolution alone. The results point to complementary information across temporal scales, with single-scale models missing part of the discriminative structure.

Dynamic fusion improves the highest-scoring two-class configuration by $2.47$ percentage points ($65.22\%$ versus $62.75\%$). For the highest-scoring three-class configuration, the difference is much smaller, with dynamic fusion exceeding concatenation by only $0.12$ percentage points ($45.43\%$ versus $45.31\%$). Across the full configuration set, the two strategies are closer; several concatenation-based settings match or exceed their dynamic counterparts. Dynamic fusion is most useful when the window set contains heterogeneous temporal scales, allowing the model to adjust each scale's contribution based on the sample.

The two tasks produce different highest-scoring temporal configurations. The two-class peak includes the full signal together with short and long windows, whereas the three-class peak excludes the full signal and combines $2$-, $5$-, and $10$-second windows. Under the present DMER evaluation, this suggests that binary discrimination benefits from both sustained and localized information, while mixed-emotion recognition benefits more from local temporal patterns.

The peak-accuracy configurations do not require the four-scale setting, which is the most computationally expensive option. Both tasks reach their best accuracy with three-scale configurations below the cost of the complete four-scale setting, although these configurations remain substantially more expensive than the full-signal baseline. The results therefore do not support using every available temporal scale.

\section{Conclusion}

This paper presented a multi-scale temporal framework for EEG-based emotion recognition. The input signal is decomposed into windows of one or several durations, encoded by a shared attention-based model, and integrated through dynamic fusion with sample-adaptive weights across temporal scales. The framework was evaluated under a subject-independent protocol for binary and three-class classification, with the three-class task explicitly including mixed affective states.

The best test accuracies reached $65.22\%$ for the two-class task and $45.43\%$ for the three-class task. Both results come from three-scale dynamic-fusion configurations and are substantially above the full-signal baseline. The highest-scoring configurations differ across tasks: the two-class peak uses the full signal along with $2$- and $10$-second windows, whereas the three-class peak uses $2$-, $5$-, and $10$-second windows without the full signal. Under the present evaluation, binary valence discrimination benefits from sustained and localized information, whereas the mixed-emotion setting appears to benefit more from local temporal patterns.

Dynamic fusion provided a clear improvement in the highest-scoring two-class configuration, while its advantage in the highest-scoring three-class configuration was marginal. Across all window configurations, the two fusion strategies performed comparably, with dynamic fusion showing its largest advantage when temporally heterogeneous scales were combined. The strongest results were obtained below the maximum-cost four-scale setting, but they still required substantially greater computation than the full-signal baseline. A compact set of complementary temporal scales is therefore sufficient for effective multi-scale modeling in this setting.

\section*{Acknowledgments}

The authors used large language model (LLM)-based tools for language editing and improvement. All scientific content, results, and conclusions are solely the work of the authors.

\bibliographystyle{IEEEtran}
\bibliography{library}

\end{document}